%% file: main.tex
\documentclass[sigconf, nonacm]{acmart}

\usepackage[noend]{algpseudocode}
\usepackage{enumitem}
\usepackage{microtype}
\usepackage{float}
\usepackage{subcaption}
\usepackage{tabularx}
\usepackage{booktabs}
\usepackage{algorithm2e}
\usepackage{multirow}
\usepackage{booktabs}
\usepackage{tablefootnote}
\usepackage{threeparttable}
\usepackage{mdframed}
\usepackage{placeins}
\usepackage{threeparttable}
\usepackage{cleveref}
\usepackage{balance}

\renewcommand{\baselinestretch}{0.99}

\usepackage{amsmath,amssymb}
\usepackage{pifont}
\usepackage{xcolor}
\usepackage{listings}
\input{code}

\definecolor{pastelyellow}{RGB}{255, 255, 230}
\definecolor{lightorange}{RGB}{255, 223, 186}
\definecolor{orange}{RGB}{255, 140, 0}

\AtBeginDocument{%
  }

\setcopyright{acmlicensed}
\copyrightyear{2018}
\acmYear{2018}
\acmDOI{XXXXXXX.XXXXXXX}
\acmConference[Conference acronym 'XX]{Make sure to enter the correct
  conference title from your rights confirmation email}{June 03--05,
  2018}{Woodstock, NY}
\acmISBN{978-1-4503-XXXX-X/2018/06}

\begin{document}



\title[Supporting Our AI Overlords: Redesigning Data Systems to be Agent-First]{Supporting Our AI Overlords: \\Redesigning Data Systems to be Agent-First}


\author{Shu Liu, Soujanya Ponnapalli, Shreya Shankar, Sepanta Zeighami, Alan Zhu \\ Shubham Agarwal, Ruiqi Chen, Samion Suwito, Shuo Yuan, Ion Stoica, Matei Zaharia \\ Alvin Cheung, Natacha Crooks, Joseph E. Gonzalez, Aditya G. Parameswaran
\\ University of California, Berkeley }

\renewcommand{\shortauthors}{Liu et al.}
\Crefname{section}{Sec.}{Sec.}

\newif\ifcomments
\ifcomments
    \providecommand{\shu}[1]{{\color{magenta}{/* shu: #1 */}}}
    \providecommand{\matei}[1]{{\color{blue}{/* matei: #1 */}}}
    \providecommand{\nc}[1]{{\color{cyan}{/* natacha: #1 */}}}
    \providecommand{\agp}[1]{{\color{red}{/* aditya: #1 */}}}
    \providecommand{\sa}[1]{{\color{brown}{/* shubham: #1 */}}}
    \providecommand{\alvin}[1]{{\color{brown}{/* alvin: #1 */}}}
    \providecommand{\souj}[1]{{\color{purple}{/* souj: #1 */}}}     
    \providecommand{\sep}[1]{{\color{magenta}{/* sep: #1 */}}} 
    \providecommand{\shreya}[1]{{\color{purple}{/* shreya: #1 */}}} \providecommand{\joey}[1]{{\color{orange}{/* joey: #1 */}}}    
    \providecommand{\alan}[1]{{\color{teal}{/* alan: #1 */}}}
    \providecommand{\shsa}[1]{{\color{teal}{/* shuo samion: #1 */}}}
\else
    \providecommand{\shu}[1]{}
    \providecommand{\matei}[1]{}
    \providecommand{\nc}[1]{}
    \providecommand{\agp}[1]{}
    \providecommand{\sa}[1]{}
    \providecommand{\alvin}[1]{}
    \providecommand{\souj}[1]{}
    \providecommand{\sep}[1]{}
    \providecommand{\shreya}[1]{}
    \providecommand{\joey}[1]{}
    \providecommand{\alan}[1]{}
    \providecommand{\shsa}[1]{}
\fi

\newif\ifgap
\gaptrue
\ifgap
    \providecommand{\gap}[1]{{\color{purple}{/* gap: #1 */}}}
    \providecommand{\sid}[1]{{\color{red}{/* Sid: #1 */}}}
\else
    \providecommand{\gap}[1]{}
\fi

\newif\ifrev
\ifrev
    \providecommand{\rev}[1]{{\color{blue}{#1}}}

\else
    \providecommand{\rev}[1]{{\color{black}{#1}}}

\fi



\newcommand{\heading}[1] {{\hfill\break\noindent{\textbf{\emph{#1}}} }}
\newcommand{\topic}[1] {{\smallskip\noindent{\textbf{#1}} }}

\newcommand{\comma}[1] {{\textit{,\space\space}{#1} }}

\input{Sections/Abstract}
\maketitle

\input{Sections/Introduction}

\input{Sections/Motivation}

\input{Sections/Architecture}

\input{Sections/Interface}

\input{Sections/QueryOpt}
\input{Sections/Storage}
\input{Sections/Conclusion}






\bibliographystyle{ACM-Reference-Format}
\bibliography{main}

\end{document}

%% file: code.tex
\newcommand{\focusTextBox}[2]{%
  { \vspace{0.5em}
  \noindent
  {\setlength{\fboxsep}{4pt}%
  \fcolorbox{darkblue}{lightbluebg}{%
    \parbox{\dimexpr\linewidth - 2\fboxsep - 2\fboxrule}{%
      \textbf{\textcolor{darkblue}{#1}}\\
      #2
    }%
  }}
  }
}

\definecolor{darkblue}{HTML}{003282}
\definecolor{lightbluebg}{RGB}{245, 248, 255}

\definecolor{codegreen}{rgb}{0,0.6,0}
\definecolor{codegray}{rgb}{0.5,0.5,0.5}

\definecolor{backcolour}{RGB}{245,248,250}
\definecolor{emph}{RGB}{166,88,53}
\definecolor{nightblue}{RGB}{9,49,105}
\definecolor{keywords}{RGB}{207,33,46}
\definecolor{lightpurple}{RGB}{130,81,223}

\lstdefinestyle{mystyle}{
    backgroundcolor=\color{backcolour},   
    commentstyle=\color{codegreen},
    keywordstyle=\color{keywords},
    stringstyle=\color{nightblue},
    basicstyle=\ttfamily\footnotesize,
    breakatwhitespace=false,         
    breaklines=true,                 
    captionpos=b,                    
    keepspaces=true,                 
    numberstyle=\small\color{codegray},
    numbers=left,                    
    numbersep=5pt,   
    xleftmargin=0.2cm,
    aboveskip=0.2cm,
    belowskip=0.1cm,
    showspaces=false,                
    showstringspaces=false,
    showtabs=false,                  
    tabsize=2,
    frame=shadowbox,
    emph={},
    emphstyle={\color{lightpurple}},
}

%% file: Sections/Abstract.tex
\begin{abstract}
Large Language Model (LLM) agents, acting
on their users' behalf to manipulate
and analyze data,
are likely to become the dominant workload
for data systems in the future.
When working with data, agents employ
a high-throughput process 
of exploration and solution formulation
for the given task, one we call
{\em agentic speculation}.
The sheer
volume and inefficiencies
of agentic speculation can pose challenges for 
present-day data systems.
We argue that data systems need to 
adapt to more natively support agentic workloads.
We take advantage of the characteristics
of agentic speculation that we identify, i.e., scale, heterogeneity,
redundancy, and steerability---to outline
a number of new research opportunities
for a new agent-first data systems architecture,
ranging from new query interfaces, to new query processing techniques, to 
new agentic memory stores. 

\end{abstract}

%% file: Sections/Introduction.tex

\section{Introduction}
\label{sec:outline}

Powered by Large Language Models (LLMs)
that can reason, invoke tools,
author code, 
and communicate with each other,
we are on the precipice of a new agentic
revolution that will transform how data systems are used.
Modern LLMs are far more {\em efficient} internally, 
matching the capabilities 
of those orders of magnitude 
larger just a year ago, 
and growing ever more {\em effective}
at understanding and manipulating 
both structured and unstructured data.
As they become both cheap and capable,
future LLM agents
will act on users' behalf: 
extracting, analyzing, transforming, and updating data---potentially 
becoming the dominant workload for data systems.

While LLM agents may match 
human reasoning capabilities, 
they won't possess 
{\em grounding}---an awareness of 
the underlying data and characteristics
of the data systems on which the data is stored. 
However, they
can make up for this lack of grounding
by tirelessly working through possible
solutions to a given data transformation task, 
far more than any human could or would.
Each individual LLM agent can theoretically issue
hundreds or thousands of requests
a second\footnote{https://developer.nvidia.com/deep-learning-performance-training-inference/ai-inference},
with this rate scaling with
the number of LLM agents. 
Many of these requests are not
attempts at a solution,
but are instead 
part of an exploratory process of metadata discovery (e.g., table schemas, column statistics),
coupled with partial solutions and validation. 
We refer to this combination of discovery and solution formulation as {\bf \em agentic speculation}---i.e., 
high-throughput, exploratory querying 
to identify the best course of action.

Agentic speculation represents a substantial departure
from present-day data systems workloads,
which are either more intermittent (e.g., from
humans or tools operating on their
behalf) or more targeted (e.g., from end-user applications). 
Consider an army of LLM
agents tasked with finding reasons
for why profits in coffee bean sales in Berkeley was
low this year relative to last. 
Since they
are not limited
by human cognitive bandwidth and
response times,
an army of agents could employ
an enormous volume
of queries to data systems, far
more than any human could---all for a single task.
Many of these queries are likely
wasteful, 
and are simply providing
the agents grounding.
As another example, if an LLM agent
is tasked with identifying 
a new crew
for a delayed flight,
it would need to 
consider various hypothetical
transactions to surface 
to a human decision maker, 
each with dozens
of updates to various 
databases.\footnote{Example thanks to Keshav Murthy at Couchbase.}
For such tasks, agents may explore many alternatives in parallel by forking database state, running speculative updates, and rolling back branches.
Overall, as agentic workloads
become more and more prevalent, 
the sheer scale and inefficiencies of agentic speculation
will become the bottleneck, and 
our data systems will need to evolve in response.

So we ask the question: {\em how can data systems
evolve to better support agentic workloads? } 
In particular, can data systems
natively---and efficiently---support agentic speculation, helping LLM
agents determine the best course of action?
This question---which, as we argue, our community
is well-equipped to answer---holds the key to
unlocking unimaginable productivity gains from 
agents being the primary mechanism 
we use to interact with data.

Thankfully, while agentic speculation represents
a new challenge for data systems,
its characteristics present new opportunites
for the redesign of data systems.
As we show, agentic speculation:

\smallskip
\noindent (1) can be {\em high throughput},
benefiting from a lot of requests to
the backend systems, issued in sequence and/or in parallel, to determine how to solve the given task.

\smallskip
\noindent (2) is {\em heterogeneous}, spanning coarse-grained 
    data and metadata exploration, partial and complete solution formulation, and validation---allowing LLM agents to make progress with approximate or incomplete outputs in early stages.
    
\smallskip
\noindent (3) has {\em redundancy}: many requests may
    access similar data or perform overlapping operations, 
    offering opportunities 
    to share computation or eliminate redundant work.

\smallskip
\noindent (4) is {\em steerable}: since speculation is fundamentally exploratory,
    if we move beyond the query-answer paradigm and allow data
    systems to more directly communicate with LLM agents, 
    it could help steer LLM requests toward the most promising directions.

\smallskip
\noindent
In this paper, we propose a new research vision for
our community 
around redesigning data systems for agents, by leveraging 
the aforementioned characteristics of speculation---scale, heterogeneity,
redundancy, and steerability.
In~\Cref{sec:motivation},
we illustrate through case studies the characteristics
of present-day agentic speculation.
In~\Cref{sec:architecture},
we propose a new architecture 
for agent-first data systems.
In Sec.~\ref{sec:interface}, \ref{sec:queryopt}, and~\ref{sec:storage},
we identify new research opportunities in
the interface, query processing,
and storage layers, respectively.

%% file: Sections/Motivation.tex

\section{Case Studies}\label{sec:motivation}

In this section we explore
the characteristics
of agentic workloads through two case studies---and identify patterns in these queries
that present optimization opportunities. 
While these case studies are simple,
they are easier to evaluate for correctness.


\input{figures/pass_and_multiturn}

We use the BIRD text2SQL benchmark \cite{bird} in our first study.
We wanted to explore if 
present-day LLMs benefit from
increasing the number of requests---in parallel or in sequence.
We used DuckDB as our
backend, and GPT-4o-mini and Qwen2.5-Coder-7B-Instruct
as two LLMs.
To first evaluate parallel requests, 
we simulated the behavior of an LLM agent ``in charge,''
with a number of ``field'' agents
each independently attempting the task, followed by 
the agent-in-charge picking one among the corresponding 
solutions. 
We plot the average success rate 
versus
the number of LLM attempts in \Cref{figure-best-of-k}. 
To instead evaluate sequential requests, we
had a single LLM agent issue queries until it was satisfied
and once again plot
the success rate versus 
the number of steps taken in \Cref{figure-multiturn-accuracy}.
We find that:

\focusTextBox{Agentic speculation---in sequence or in parallel---helps improve accuracy.}{The success rate 
of agentic workloads
increases as a function of requests, and by 14\%--70\% in our case study.}

\input{figures/unique_subexpr_op_size}

\noindent Next, we quantify the degree to which work sharing is possible across requests.
We focus our attention on the parallel setting, 
with 50 independent attempts---and 
evaluate the redundancy across these attempts.
We plot the total number and distinct number of sub-plans or sub-expressions of each size in the 50 query plans generated for a given task,
aggregated across the full BIRD dataset,
in \Cref{figure-unique-subexpr-size}. 
We present a similar plot for sub-plans
grouped by root operator type in \Cref{figure-unique-subexpr-op}.
We find:

\focusTextBox{Agentic speculation has substantial redundancy across requests.}{Across queries,
the number of distinct sub-plans of each size is often 
a small fraction of less than 10-20\% of the total, representing
considerable potential for sharing computation.}

\begin{figure}
    \centering
    \includegraphics[width=0.9\linewidth]{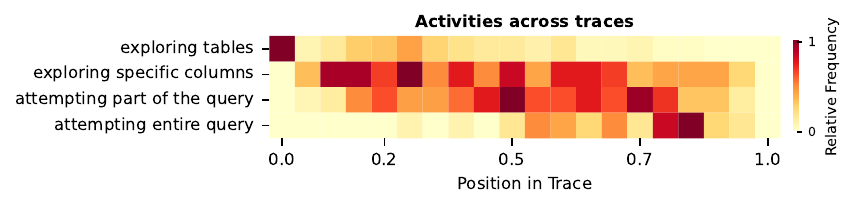}
    \vspace{-15pt}
    \caption{Labeled agent activities, with x-axis showing normalized position in the trace, and each row (activity) normalized independently. Agents first explore table and columns then formulate queries, with phases often overlapping.
    }
    \label{fig:trace_heatmap}
    \vspace{-10pt}
\end{figure}

Our second case study is more involved
than text2SQL and helps us
study the phases of agentic speculation. 
We evaluate the performance of a data agent that must combine information from two separate backend databases, chosen from PostgreSQL, SQLite, MongoDB, and DuckDB.
For example, one task involves cleaning customer information from MongoDB to join with user interaction data (e.g., upvotes) in DuckDB.
As such, it is impossible to complete
this task in a single shot,
and successful attempts
typically involve interacting with both backends,
followed by some computation in Python.
We collected 44 sequential traces of OpenAI's o3 model
attempting each of the 22 tasks twice,
with about half resulting in correct answers.
We then \textit{manually} labeled each action taken 
by the LLM with an annotation:
exploring metadata and sample data
(targeting schemas or with \texttt{LIMIT}),
exploring column statistics (distinct values or aggregates),
attempting part of the query, or all of it.
As we can see in the aggregated heatmap of
traces in \Cref{fig:trace_heatmap},
exploring metadata and sample data typically happens first,
followed by statistics, 
after which the next two phases emerge.
However, these phases are not clearly
delineated, and each phase is present
throughout the trace.
So we find:

\focusTextBox{Agentic speculation is heterogeneous in its information needs.}{Requests from agents vary greatly
in the information necessary, from coarse-grained exploration of metadata
and data statistics, to partial or more complete attempts at addressing the task.
Coarse-grained, exploratory
requests typically happen early on.}

\noindent In the following, 
we describe the earlier phases
as {\em metadata exploration},
and the latter phases
as {\em solution formulation}.


\begin{table}
\centering
\footnotesize
\caption{Mean activity counts per agent trace, averaged across all traces, with and without human expert-provided hints.}
\vspace{-10pt}
\begin{tabular}{@{}lrrr@{}}
\toprule
Activity & Avg (No Hints) & Avg (w/ Hints) & Reduction (\%) \\
\midrule
exploring tables             & 3.44 & 2.95 & -14.2 \\
exploring specific columns   & 3.56 & 2.57 & -27.7 \\
attempting part of the query & 4.28 & 2.71 & -36.6 \\
attempting entire query      & 1.26 & 1.05 & -16.6 \\
all SQL queries                   & 12.67 & 10.38 & -18.1 \\
\bottomrule
\end{tabular}
\label{tab:activity_counts}
 \vspace{-10pt}
\end{table}

Next, 
we wanted to explore if grounding
provided by the backend system could help reduce
the number of steps taken to reach the solution.
So, we simulated this by 
measuring the impact of injecting hints into 
the prompt, where the hint provides background
information useful for the task, such as 
which column contains information pertinent to the task.
Again, we collected 44 sequential traces (two per task) with hints provided,
and then measured the average number of steps
required across attempts and tasks
when hints were provided versus not.
As shown in \Cref{tab:activity_counts}, 
the impact of hints is substantial. We find:

\focusTextBox{Agentic speculation is steerable through grounding hints.}{Speculation traces can become much more efficient---reducing
queries by $>$20\%, depending on phase---if proactively
provided grounding pertinent to the task.}

Based on the characteristics gleaned
via our case studies, we
next propose a new architecture
for agent-first data systems. 

%% file: figures/pass_and_multiturn.tex
\begin{figure}[hbtp]
  \vskip -0.1in
  \centering
  \begin{subfigure}[b]{0.45\linewidth}
    \includegraphics[width=\linewidth]{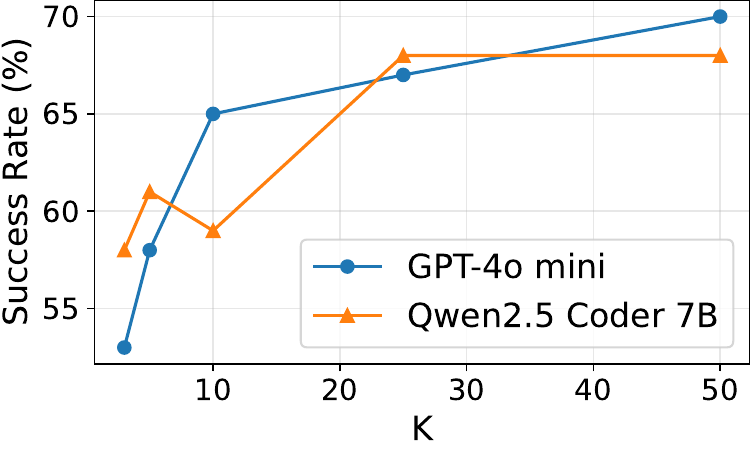}
    \caption{Success @ K}
    \label{figure-best-of-k}
  \end{subfigure}
  \hfill
  \begin{subfigure}[b]{0.45\linewidth}
    \includegraphics[width=\linewidth]{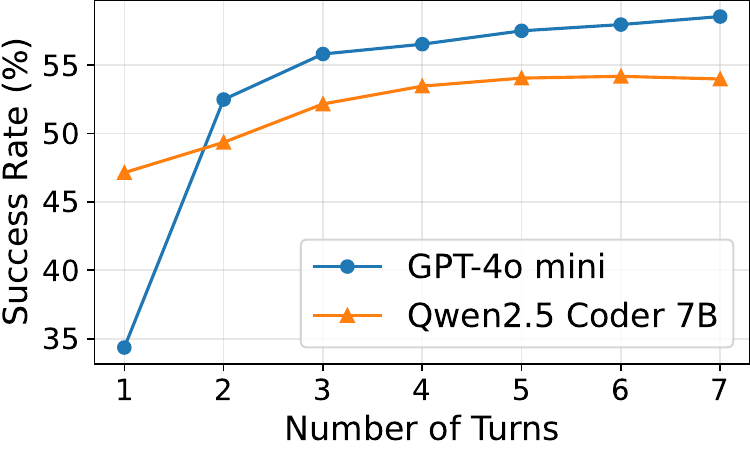}
    \caption{Success vs. Turns}
    \label{figure-multiturn-accuracy}
  \end{subfigure}
  \vspace{-10pt}
  \caption{Results on the BIRD dataset
  }
  \label{fig:bird-overall}
  \vskip -0.15in
\end{figure}

%% file: figures/unique_subexpr_op_size.tex
\begin{figure}[hbtp]
  \vskip -0.1in
  \centering
  \begin{subfigure}[T]{0.45\linewidth}
    \includegraphics[width=\linewidth]
    {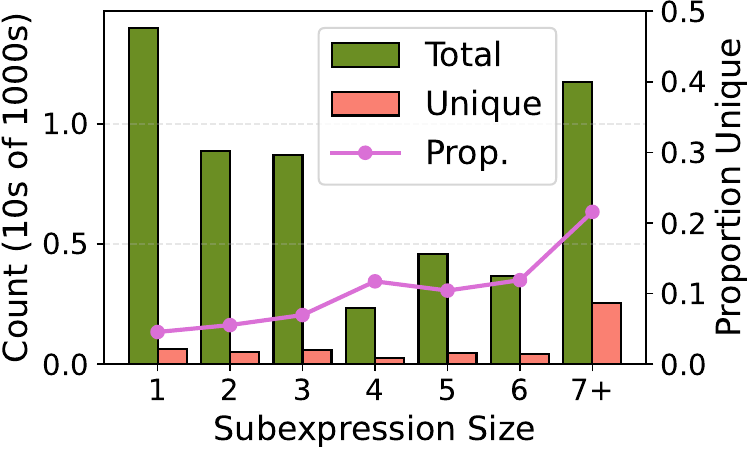}
\caption{versus subexpression size.}
\label{figure-unique-subexpr-size}
  \end{subfigure}
  \hfill
  \begin{subfigure}[T]{0.45\linewidth}
    \includegraphics[width=\linewidth]
    {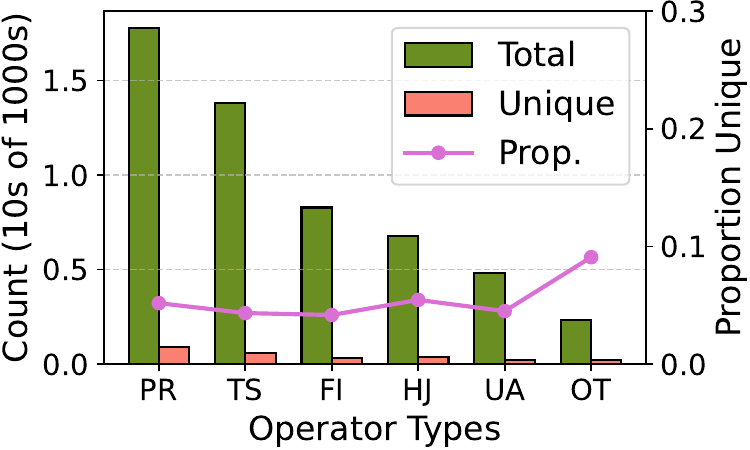}
\caption{versus root operation.}
\label{figure-unique-subexpr-op}
  \end{subfigure}
  \vspace{-10pt}
  \caption{Total vs.~unique subexpressions (count and proportion) across 50 attempts generated by GPT-4o-mini per problem, aggregated over the full BIRD dataset. Here, 
  PR=Projection, TS=Scan, FI=Filter, HJ=Hash Join, UA=Aggregate, OT=other operations. 
  }
  \label{fig:bird-subexpr}
  \vskip -0.15in
\end{figure}

%% file: Sections/Architecture.tex
\section{Agent-First Data System Architecture }\label{sec:architecture}
\begin{figure}[hbtp]
\vskip -0.4in
\begin{center}
\centerline{\includegraphics[width=0.8\linewidth]{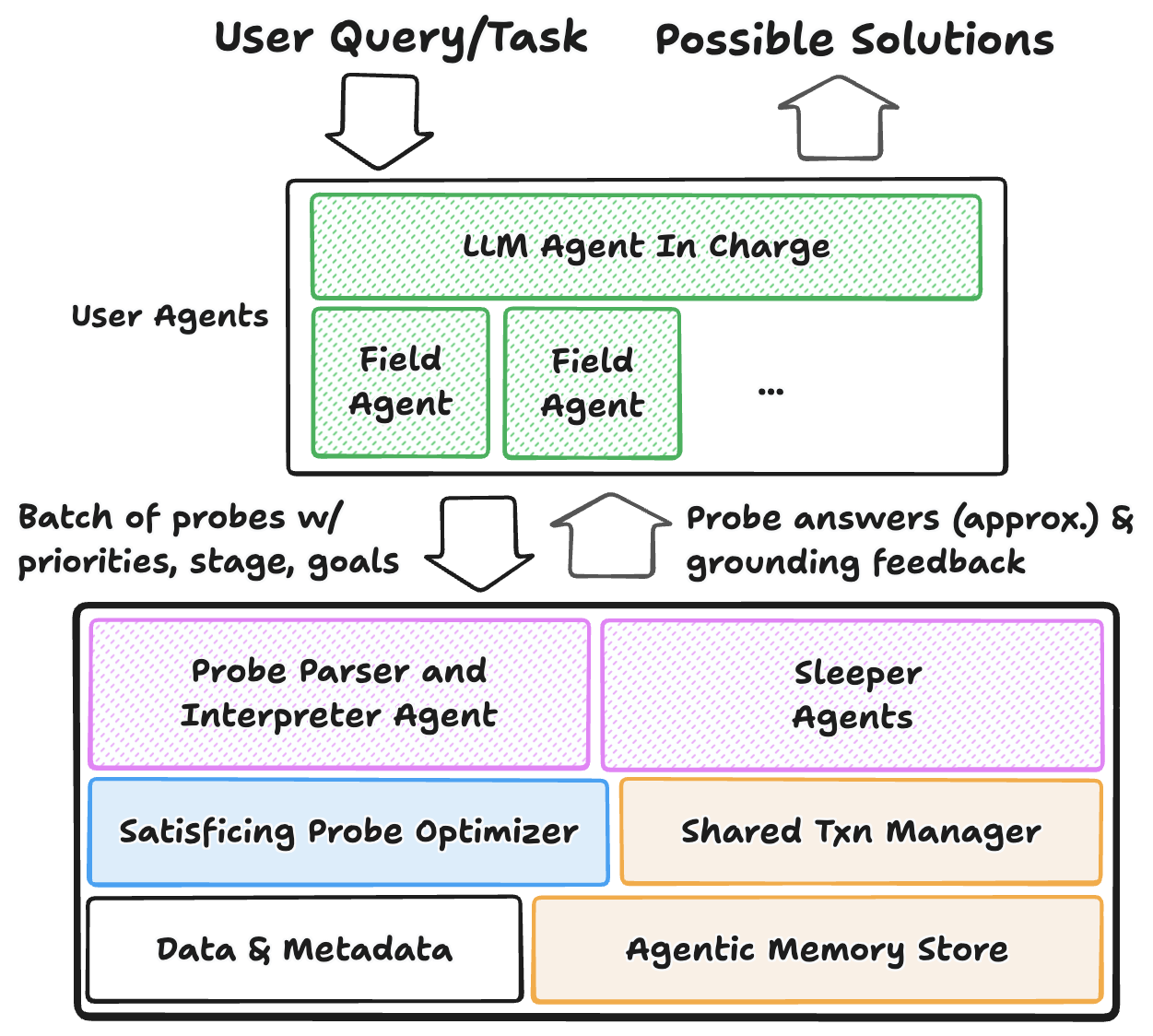}}
\vskip -0.15in
\caption{Agent-First Data Systems Architecture;
components that are dashed involve LLM agents.
Boxes in pink are covered in \Cref{sec:interface};
blue in \Cref{sec:queryopt};
orange in \Cref{sec:storage}. 
}
\label{fig:arch}
\end{center}
\vskip -0.3in
\end{figure}

Here, we outline a potential
architecture for a 
data system that is {\em agent-first},
as shown in \Cref{fig:arch}. 
\matei{"probes" are a nice new concept, but they are not well-defined -- I'd love to see an example; we don't mention them in the intro and don't really highlight them in the diagram or in the formatting of the text (e.g. there's no subsection or heading on them), so it somehow looks like we're minimizing this concept. It might be nicer if we somehow said up front that we propose a design with 2-3 new concepts, which are probes, agentic memory and whatever, and put those in bold and made them really stand out somehow as you read the paper.}

Given a user task,
an army of LLM agents
can issue one or more {\bf \em probes}
to the backend system, possibly
associated with relative priorities.
We call these {\em probes} rather than queries
for two reasons. 
First, they could go beyond SQL
in providing
background information
about the nature of the request,
such as the phase (metadata exploration
or solution formulation),
the identity of the agent issuing the request,
the degree of accuracy required, 
overall goals, among others.
We envision this information to be specified
in natural language or some 
other flexible format
to be interpreted by in-database agents.
Second, the probes could go beyond SQL
on data or metadata (e.g., via \texttt{information\_schema}) to 
search for tokens that may be present in any table 
(either column or row)
to help identify which tables need
to be accessed.

Then, these probes are parsed and interpreted
by an agentic interpreter component 
within the database. 
For each of these probes, the system could
provide answers, possibly approximate, 
and also proactively
provide information going beyond answers
to help {\em steer} the agents
through grounding feedback. 
We describe our interface
as well as proactive feedback in \Cref{sec:interface}.

Given one or more probes,
our probe optimizer
attempts to {\bf \em satisfice}, i.e., 
produce reasonable 
results that address needs, without evaluating the query completely,
as described in \Cref{sec:queryopt};
this optimizer
leverages and extends traditional
database research
on multi-query optimization
and approximate query processing.

To improve efficiencies,
the storage and transactional
components of our data systems
will need to evolve,
as described in \Cref{sec:storage}.
We introduce an {\bf \em agentic memory store} 
to store any grounding
gleaned, so that they can 
be used in future probes.
For updates, 
our {\bf \em shared
transaction manager}
efficiently handles the sheer redundancy
in state involved across 
many potential transactions. 

%% file: Sections/Interface.tex
\section{Query Interfaces}\label{sec:interface}
In this section, 
we focus on {\em agent-database interaction}.
We start by describing how probes (i.e., input from agents
to the data system) need to go beyond SQL in \Cref{sec:interface:input}.
Then, we discuss how 
data systems can go beyond the query-result
paradigm in providing additional grounding
information to help steer the 
agents in \Cref{sec:interface:output}.



\subsection{From Agents to Data Systems}\label{sec:interface:input}
Probes from agents need to go beyond SQL
in specifying {\em why} or {\em how}
a given query needs to be answered.
Moreover, for certain types of information needs,
SQL may be limiting, necessitating
 new operators. 
We describe each aspect in turn.


\topic{Providing Background Information.} 
If all an agent can do is specify SQL queries,
then all the data system can do is provide exact
results for those queries,
making speculative probing inefficient.
While specifying \texttt{LIMIT}
or exact degree of approximation is
one option, it provides limited expressive
power.
Therefore, as part of a probe,
agents can specify one or
more SQL queries, along with what
we call a {\em brief}:
natural language statements 
about the probe's goals and intents,
its current phase (metadata
exploration or solution formulation),
approximation needs and priorities across queries or probes,
as well as any other open-ended information.
These briefs
are in turn examined by the probe interpreter
agent within the data system and used to guide 
optimization and execution,
e.g., what order to execute the queries (if at all)
and degree of approximation (or accuracy) based on goals and phase.
Determining how to set accuracy based
on this natural language input is an open question
and needs to also take into account
relative query execution costs.

Across a batch of queries specified 
within 
a probe,
the probe can additionally specify
open-ended goals that go beyond simple accuracy,
such as pair-wise priorities,
or indicating that only $k$ query among 
$n$ specific queries needs to be performed to 
completion (and the data system can decide which one to maximize efficiency).
For example, if a field agent
in an exploratory phase
wants to get a sense for the differences
in sales performance of stores on the US West
coast vs. East coast,
it can specify, as a part of the probe,
that the data system needs to generate statistics
for two states each from each coast,
with the data system being able to pick which ones.
The interface can furthermore 
allow for other forms of approximation 
indicators that are time-consuming 
for humans to write but can now be done by agents, 
e.g., specifying \textit{termination criteria}, 
functions that the data system can evaluate
on the partial result sets to know 
if some queries can be terminated early. 
For example, one termination criteria 
could be defined to stop execution 
of multiple ``needle-in-a-haystack'' 
type queries mid-execution
because the answers are too similar to 
previous ones  (where an agent defined function is evaluated on partial result sets to determine answer similarities). 

\topic{Extending Capabilities through Flexible Probes.} 
In many cases, agents are unsure of even where
to start and which tables to query for a given
task---because
they lack knowledge of how the data is organized.
Suppose an agent is tasked with finding out
how a given company will be ``impacted by increased tariffs
on the import of electronic goods.''
This agent may want to find tables 
whose name is semantically similar to ``electronics,''
or whose rows contain data that is semantically similar.
Such probes
that ask for semantically similar
contents---be it tables, columns, or rows---to a specific phrase, located {\em anywhere},
are impossible to address within SQL,
but are valuable during the early exploratory phases.
Thus, we need native support for semantic similarity 
operators,
beyond \texttt{LIKE},
where the operators are applied to any data
or metadata in the data system.
Furthermore, 
as we will discuss 
in \Cref{sec:storage},
the agents will rely on metadata 
stored in agentic memory,
on cells, rows, columns, and tables, 
typically written by agents themselves---to 
understand data semantics, 
and as such will need to frequently 
query or update this metadata. 
Although the above functionalities 
may be possible through a combination of tools 
(e.g., store metadata separately in a vector database, 
look it up and then issue SQL queries), determining
what and how to actually store, and how to keep it up-to-date
is a challenge. Moreover, a data system that holistically 
supports all data and metadata needs can be more 
effectively used by agents. \alvin{can we differentiate this from today's vector db and RAG?} \shreya{I have a comment similar to Alvin's---but, if we don't really have anything new to say about the importance of semantic search and metadata, we could potentially shrink much of the paragraph} \souj{Agreed, or move some of the metadata store discussion to S6!}

\subsection{From Data Systems to Agents}\label{sec:interface:output}
In addition to simply
answering probes, 
data systems should \textit{steer} 
agents towards better probes,
which in turn can lead to improved efficiencies. 
In this way, the data system
acts in a more \textit{proactive} \cite{zeighami2025llm} manner,
akin to how a
data engineer or administrator
may assist data analysts
in satisfying information needs as efficiently as possible.
This information can be provided
in addition to, or in lieu of the
answers to the probes, in natural language. 
This steering can serve two purposes: 
(1) helping agents by providing auxiliary data-centric information 
the data system 
finds relevant, as a side-channel, 
and (2) providing feedback to agents 
on efficiency and costs to assist the agents in designing their probes. 
We envision {\em sleeper agents} within the data system
that are invoked on-demand
to gather information in parallel with answering probes, 
to be returned in addition to probe answers, 
as we discuss below. 

\topic{Auxiliary Information.}
As we saw in \Cref{tab:activity_counts},
providing grounding hints or feedback can reduce
the number of probes agents need
to complete their tasks.
We envision sleeper agents
tasked with identifying and providing such hints
as auxiliary information along with answers.
For example, the sleeper agent could find
and share
other related tables---to be either joined with (as in
join discovery, e.g., \cite{sarma2012finding}), or
replacing the current table
as the focus of analysis, especially if 
the current table proves irrelevant.
Or rather than the agent having to guess
why they got an empty result, the 
sleeper agent can provide feedback reminiscent
of why-not provenance~\cite{cheney2009provenance},
e.g., the probe assumed that states
were encoded with two letter acronyms like ``CA'', but
instead they are listed out in entirety.


\topic{Cost Estimates and Cost-Based Feedback.} 
Grounding can also come in the form
of cost estimates; for example, even before
executing a query, 
estimated costs (especially if higher than expected)
can be provided to the agent to help determine
if the probe must be run to completion,
and suggest the agent to modify the probe (e.g.,
to just focus on California instead of all of USA),
or increase the degree of approximation. 
This can similarly be applied across probes. 
For example, 
if the sleeper agent predicts
that the probes are performing a set of tasks in sequence,
it can suggest to the field agent to batch them,
if it proves to be cheaper.
The sleeper agent can also take into account
related materialized answers, or if a similar query was just answered for another agent. In such cases, the sleeper agent can suggest modifying the input probe to probes with such pre-defined answers to improve efficiency---or it can output the answer for such related probes in the side-channel. 

Next, we discuss how to efficiently provide answers to probes.

%% file: Sections/QueryOpt.tex
\section{Processing and Optimizing Probes}
\label{sec:queryopt}


As discussed in~\Cref{sec:motivation}, agentic probes will have much higher throughput than those issued by human sources (e.g., web applications).
Importantly, in agent-first data systems, our goal is not to optimize overall throughput as in traditional databases, but to {\em evaluate probes enough such that agents can make their decision on how to proceed in the next turn.} With that in mind, this section discusses what needs to change in data systems to effectively support probes.

\subsection{Supporting Exploration}
Our agentic probes will consist of exploratory queries to establish grounding. Some explorations will inevitably be cast in natural language (NL) as agents may lack knowledge about the underlying databases (e.g., ``how to find out how many tables are stored?'') with others expressed using SQL (e.g., {\small {\tt SELECT count(*) FROM information\_schema.tables}} in PostgreSQL). Today's databases are not designed to answer NL queries. The probe optimizer in our agent-first data system must therefore orchestrate the mix of NL and SQL queries by utilizing different agents at the scale of probes.


To illustrate, consider identifying the stores that show an increasing sales trend. Our agents will first need to find out which tables are used to store sales data. A straw person probe execution plan is to pose NL questions to a web search agent to discover how to look up table schema for our specific database dialect, and execute the found queries on our database. While these are simple queries on our database's metadata tables, the outputs returned from such queries often contain lots of unnecessary information. For instance, PostgreSQL maintains hundreds of internal tables and indexes even without any user table defined. Coupled with the user tables, the results can easily grow to thousands---or hundreds of thousands---of rows. Feeding all the rows to our query formulation agent is a waste of its limited context length. As mentioned in Section~\ref{sec:interface:input}, we further need the ability to query tokens regardless of where they appear across databases, be it as part of
metadata or data. 

Subsequently, to discover what constitutes an increasing trend, one strategy is to find examples of ``trend queries'' (possibly using window queries) using NL with a web search agent, then feed the returned information to a query formulation agent to translate into SQL. We will likely get lots of example queries online, and our  database will be bombarded with lots of inapplicable queries (e.g., they refer to non-existent tables, 
or identify the wrong trend). Worse yet, all such explorations will be mixed with other agents formulating solutions. With today's data systems, we have no means to identify which queries are part of agentic exploration (and hence do not need to be evaluated completely). We envision that our probe optimizer will prioritize queries based on their phases (i.e., a form of admission control). Furthermore, we will store previously gleaned information using our agentic memory store to avoid repeated querying of the same information, and train agents to query our memory store instead of including such information as part of the prompt each time.




\subsection{Probe Optimization}
As mentioned, probes issued by agents, unlike queries issued by humans, do not require complete answers. The database interface allows the agents to specify goals, and approximation needs in natural language via \textit{briefs}, which are then used by the database to decide which probes to execute and to what degree of accuracy. This means the goal of the query optimizer, unlike in traditional data systems, is to decide both \textit{what} queries to execute (and to what degree of approximation) to \textit{satisfice} for the probe, as well as \textit{how} to execute them. In doing so, the optimization has a new objective: \textit{minimize the total time spent on answering the field agents' probes given available computational resources.} Solving this optimization problem requires the database internally balancing cost/accuracy trade-offs: if the database chooses to answer a query with high degree of approximation providing insufficient answers to save cost upfront, the agent may ask many follow-ups with increased accuracy requirements, thus increasing total time spent answering the agent's probes. We next discuss how we envision such an optimization problem can be solved, within a given batch of probes sampled at an interaction turn (\Cref{sec:optimization:intra}), and across batches of probes across turns and agents (\Cref{sec:optimization:inter}).  


\subsubsection{Intra-Probe Optimization}\label{sec:optimization:intra}
We first discuss how to optimize a given batch of probes to provide sufficient information for the agent while minimizing computational cost.  

\topic{Deciding What to Execute.} The database must first decide what queries to run and to what degree of approximation, taking the probe and its briefs into account. This requires the database to reason about the data and probe semantics, including the agent's goals and phases.
To do so, the database can use semantic query and data understanding to check if they match user's intents, and prune away queries it deems not semantically meaningful. For instance, during the exploration phase, the database can examine the projected columns in probe to see if they are relevant to the user's intent, and if not prune such columns, or the probes away as a whole. Moreover, the database can compare probes within a batch, guided by probe's briefs that may have specified approximation needs across probes. The database can then make cost estimates and compare information gain from the probes to decide which probes are more helpful and/or cheaper. For example, given two probes $P$ and $P'$ the database can prune $P'$ away if rows that would be returned by $P' - P$ are deemed irrelevant to the agent's goal. This is reminiscent of prior work on pruning queries as part of visualization recommendation~\cite{seedb}, and deciding query equivalence as part of query synthesis given user provided input/output examples~\cite{zloof, scythe1}, although the scale of queries to compute the differences will be much larger in agentic workloads. Finally, the database can take the agent's phase into account; for example, return coarse grain approximations during exploration, but more accurate answers during solution formulation. Beyond pruning queries, we envision agents will be able to examine other internal database states (e.g., buffer pool, outputs of query operators) to determine if it should continue with query evaluation, or move on to the next turn.


\topic{Efficient Execution.} 
As mentioned in Sec.~\ref{sec:motivation}, probes have substantial redundancy that we can exploit by sharing computation across them. Multi-query optimization~\cite{sharedb, mqo1, mqo2}, approximate query processing~\cite{aqp} and caching of partial query results can be used to improve efficiency. However, there are new unique challenges. For example, different probes will have different approximation requirements and may be accompanied with termination criteria (a function that can be evaluated on partial results to know if they are sufficient, see Sec.~\ref{sec:interface}), which makes it more difficult to reason about their semantics and what can be shared. 
Moreover, the database can incrementally evaluate queries, reminiscent of incremental query processing~\cite{trill}, but with the new challenge of decision making across them; e.g., the database must decide which probe is the most useful to the agent and provide higher accuracy for that probe first before increasing accuracy for other probes. Finally, query planning and processing can be done jointly with optimization, e.g., the database can re-evaluate its decisions on what queries to run, or increase its level of approximation for some query during planning or processing as it obtains more information.

\subsubsection{Inter-Probe Optimizations}\label{sec:optimization:inter}
The database can furthermore leverage the sequential interactions with agents across turns to further optimize both the queries it decides to run and their execution.   

\topic{Deciding What to Execute.} Besides the strategies discussed in \Cref{sec:optimization:intra}, the database can consider all interactions with the agent to decide what queries to run. First, it can decide on queries to run based on whether they provide any new information given past queries answered. For example, when given probes $P$ and $P'$ by the agents across consecutive turns, if the output between $P$ and $P'$ is expected to not provide new information to the agent---e.g., $P'$ adds new columns that are non-relevant to the agent's goal---then $P'$ can be dropped. Furthermore, the database can decide what queries to run in order to minimize the number of future follow-up probes. For example, based on the agent's goal specified in the probe briefs, it can run a query it finds maximally useful to the agent exactly and to completion rather than approximately even if the current query may take longer, expecting that the extra computation upfront will reduce total runtime across future interactions with the agent. Yet another direction is to treat the problem as one
of exploration vs.~exploitation: instead of always trying to provide
rapid answers to queries by satisficing, the database can sometimes prioritize exploration of underexplored solution spaces to identify
those solutions that have an unanticipated benefit, in order to 
maximize utility over time.


\topic{Efficient Execution.} The database can decide to materialize and cache answers by observing the query history and considering the agent's intent. For example, based on the history and the agent's intent, the database can expect future probes will continue to involve the join of certain tables and can materialize the join. 

%% file: Sections/Storage.tex
\section{Indexing, Storage, and Transactions}\label{sec:storage}



The heterogeneity and redundancy
of agentic speculation workloads 
fundamentally challenge the assumptions
of the storage layer of 
today's data stores,
specifically, that workloads
are static and independent.

For static workloads,
data systems rely on 
predefined indexes and fixed storage layouts 
(e.g., column-based for OLAP) 
based on recurring workload patterns.
Agentic probes, by contrast, evolve from coarse-grained metadata exploration to final validation. 
This dynamism makes static tuning ineffective. 
Meanwhile, the exploratory (resp. solution formulation) phases
of different probes may 
be similar and can benefit from similar
layouts.

On the independence front,
data systems treat queries
as unrelated, such that
concurrent access (specifically writes) from these queries
must be isolated from each other.
While this simplifies application logic 
and ensures consistency, these mechanisms prevent
cooperative sharing of state with rare exceptions~\cite{entangled}. 
Instead of isolation, agentic workloads
demand a more cooperative model---one
that can safely share intermediate
state across different probes, many of which 
are likely to be similar.

Hence, we propose two key ideas to improve performance.
First, we propose an
agentic memory store that acts as a ``pseudo-index''
to help agentic probes quickly find
information that may be helpful,
either directly accessed by them,
or on their behalf by sleeper agents.
Second, we propose a new
transactions framework that 
is centered on state sharing 
across probes,
each of which may be independently
attempting to complete
a user-defined sequence of updates.



\subsection{Agentic Memory Store}


The exploratory phase of agentic speculation
aims to identify the right 
tables and columns to operate on. 
To improve efficiency,
data systems should maintain
a persistent, 
queryable {\em agentic memory store}---a semantic
cache that provides grounding.

\topic{Artifacts.}
The first question is what should be stored.
One idea is to store
the results of prior probes and partial solutions,
so that agents can reuse what is known
about the data and metadata, enabling similar probes
to be more efficient. 
In addition, we can 
store information about the data and metadata,
possibly associated with the tables themselves.
We can store 
encoding formats for columns, 
missing value information,
and time and location granularities.
For example, an agent trying to explore
various sales partitions may retrieve
a number of them, along with the 
metadata in the agentic memory that indicates 
the date ranges or location ranges associated with each---so
that it can make a more informed decision about which ones
to probe further.

To implement this store, we can embed the agentic metadata
with the table directly, to be retrieved if the table is queried.
For all other open-ended information, one
approach is to use a vector index
to support semantic similarity search on embeddings
(e.g., querying with a probe might retrieve other similar 
probes,
and what worked for them).
However, this approach may not work as well
for more targeted or more structured lookups.

\topic{Updates to the Store.}
A separate concern is how this memory
store is maintained during updates.
Updates could be in the form of new probes
being executed that may provide new information
that augments or supersedes existing ones.
Or, it could be to the underlying data or metadata,
necessitating updates to any related information
in the agentic memory.
For example, if there is a schema update,
the results of a prior probe that used that table may
no longer be relevant.
One approach is to allow this memory store
be inconsistent with the data/metadata,
and instead be updated by any new probes that discover
that the information is stale.
However, the downside is that the stale information may
lead a new probe
to make a mistake.
For example, suppose the agentic memory indicates
that the only relevant sales information can be found
in three tables, but after that, additional relevant
tables were added; here, new probes may
end up returning incorrect results.
Additional challenges emerge
in supporting access control for multiple users.
For example, agents acting on
different users' behalf could ask similar questions (e.g., "Where is the employee’s availability stored?"). 
Sharing answers across such agents boosts efficiency---but raises privacy concerns,
especially in the aggregate~\cite{cryptdb}. 
Addressing these challenges
will need to draw inspiration from work
on knowledge bases as well as schema evolution.

\subsection{Performing Branched Updates}

When transforming or updating data,
agents typically explore multiple ``what-if''
hypotheses, i.e., branches.
For example, at Neon~\cite{neon},
we observed that
agents created 20$\times$ more branches,
and performed 50$\times$ more rollbacks,
relative to humans.
Traditional transactional guarantees
instead operate within a linear
thread of execution.
Here, with agentic speculation, we instead
want multi-world isolation,
where each branch must be logically
isolated, but may physically overlap.

\topic{Branch Isolation.} 
Existing models of branching consistency, 
developed in the weak consistency era, e.g., in Bayou, Dynamo, or Tardis~\cite{bayou,dynamo,tardis}, 
as well as versioned databases~\cite{huang2017orpheusdb}
can offer inspiration. 
However, agentic speculation goes further: multiple agents may create forks that must eventually reconcile---not just with the mainline, but with each other. This requires new models of multi-agent, multi-version isolation. Most branches will be similar---e.g., same schema, 90\% identical data---but isolation requires that their effects remain logically separate.

\topic{Efficient forking and rollbacks.} 
Naively duplicating entire data\-bases per branch is prohibitively expensive and inefficient, making support for efficient forking crucial. Industrial systems like Neon~\cite{neon}, Aurora~\cite{aurora}, and Bauplan~\cite{tagliabue2025safe} and academic systems like Tardis~\cite{tardis} adopt copy-on-write approaches to lazily clone state. 
However, these are still far from what is needed for agentic speculation at a massive scale. 
We need new concurrency mechanisms
that exploit similarity across
branches and preserve logical isolation (no cross-contamination),
to enable massive parallel forking.
This is analogous to MVCC on steroids: 
forking possibly thousands of near-identical snapshots 
and rolling back all but one. 
Unlike traditional data systems,
where rollbacks are rare, we
require ultra-fast rollbacks 
(i.e., fast aborts for failed branches).

%% file: Sections/Conclusion.tex
\section{Conclusion}
We described
our vision for data 
systems that natively support 
emerging agentic workloads. 
These
workloads involve 
agentic speculation, characterized by a
high-throughput, heterogeneous, and
redundant mix of 
discovery and validation, 
specified 
by probes ideally involving a combination 
of queries and natural language. 
We present one such architecture 
for such redesigned data systems, 
and discuss emergent research challenges.

\balance
\subsection*{Acknowledgments}
This work is supported by NSF grants IIS-1955488, IIS-2027575, DGE-2243822, IIS-2129008, IIS-1940759, and IIS-1940757, DOE awards DE-SC0016260, AC02-05CH11231, DARPA agreement HR00112\-590131, and funds from the state of California. This work is also supported by EPIC Data Lab sponsors and affiliates, including Adobe, Bridgewater, Google, G-Research, Microsoft, PromptQL, Sigma Computing, and Snowflake, as well as Berkeley Sky Lab sponsors and affiliates, including Accenture, AMD, Anyscale, Broadcom, Google, IBM, Intel, Intesa Sanpaolo, Lambda, Lightspeed, Mibura, Samsung SDS, SAP, Cisco, Microsoft and NVIDIA. Compute credits were provided by Azure, Modal, NSF (via NAIRR), and OpenAI. We thank the reviewers for their feedback, as well as Arash Nourian for helpful discussions. 